\documentclass[sigconf]{acmart}  

 \usepackage{booktabs}
 \usepackage{subfig}
 \usepackage{cleveref}
\title[Decentralised Multi-Demic EA Approach to the MATSP]{Decentralised Multi-Demic Evolutionary Approach to the Dynamic Multi-Agent Travelling Salesman Problem}

\author{Thomas Kent}
\affiliation{%
\department{Department of Computer Science}
  \institution{University of Bristol}
  \city{Bristol, UK}
}
\email{thomas.kent@bristol.ac.uk}
\author{Arthur Richards}
\affiliation{%
  \department{Department of Aerospace Engineering}
  \institution{University of Bristol}
  \city{Bristol, UK}
}
\email{arthur.richards@bristol.ac.uk}

\begin{document}

\begin{abstract}
The Travelling Salesman and its variations are some of the most well known NP hard optimisation problems. This paper looks to use both centralised and decentralised implementations of Evolutionary Algorithms (EA) to solve a dynamic variant of the Multi-Agent Travelling Salesman Problem (MATSP). The problem is dynamic, requiring an on-line solution, whereby tasks are completed during simulation with new tasks added and completed ones removed. The problem is allocating an active set of tasks to a set of agents whilst simultaneously planning the route for each agent. The allocation and routing are closely coupled parts of the same problem making it difficult to decompose, instead this paper uses multiple populations with well defined interactions to exploit the problem structure. This work attempts to align the real world implementation demands of a decentralised solution, where agents are far apart and have communication limits, to that of the structure of the multi-demic EA solution process, ultimately allowing decentralised parts of the problem to be solved `on board' agents and allow for robust communication and exchange of tasks.
\end{abstract}
\maketitle


\section{Introduction And Background}

Many real-world problems such as reconnaissance and surveillance, search and rescue and package delivery rely on decision making and coordination of multiple agents \cite{Alighanbari2004,Ramchurn2015a,Groen2007}. These can be roughly broken down into navigating to a location, completing a task and moving on to the next. The question of this paper is then, given a number of tasks to be completed and a number of agents to complete them, what is the best way to allocate tasks to agents and subsequently navigate between those tasks.

The problem can be defined as both allocating a set of tasks to a number of agents and simultaneously planning the route for each agent \cite{Korsah2013, Bektas2006a}. This is a slight variation on the classical Multi-Agent Travelling Salesman Problem (MATSP). Given $A$ salesmen (agents) and $N$ nodes (tasks) at different locations, a solution to the MATSP seeks $A$ trails such that each node (task) is visited only once and by only one salesman (agent), whilst minimizing a given cost function.

In the MATSP the allocation and routing are a closely coupled problem. Defining a route implicitly determines the allocation, whereas the inverse is not necessarily true. If all the tasks were allocated first, then calculating the remaining route for each agent would be comparatively trivial (i.e. multiple single travelling salesman problems). It is in the moving of tasks between agents that makes this problem dynamic and in turn more interesting. It is this problem structure that this paper looks to exploit through the use of Evolutionary Algorithms (EA), Multi-Demic (or Multi-Population) EA and ultimately a decentralised Multi-Demic EA. The real driving question of this work is: can the real-world constraints of the problem, such as limitations to communication, need for robustness and spatial separation of agents, inform the structuring of the optimisation technique in such a way that is mutually beneficial to both the solver and execution?



Evolutionary Algorithms, also known as genetic algorithms or genetic programming, are stochastic meta-heuristic search algorithms which have shown to be effective at solving hard optimisation problems \cite{Alba2002,Zhang2015, Louis1999,Lopes2016, Potvin1996}. Inspired by biological evolution they employ the idea of maintaining a population of individuals, which are candidate solutions, along with mechanisms for reproducing, mutating and selecting members of the population to produce new `generations'. They have been used to solve a wide range of optimisation problems due to their heuristic nature and flexible techniques.

Currently a large amount of research effort is focused on implementing both parallel and distributed EA variations. Zhang et. al \cite{Zhang2015} explore the state of the art of current distributed evolutionary algorithms, comparing and contrasting a wide range of structuring both the populations and the dimensions of problems. Importantly they show that the distributed nature of the population allows a Distributed Evolutionary Algorithm (DEA) to maintain a diverse population, potentially avoiding local optima. The work of Alba and Tomassini \cite{Alba2002} looks at the range of parallelization techniques used for EAs and some of the associated algorithmic issues. Sarma \cite{Sarma1998} notes that an important operator in EA, selection, can be implemented in a decentralised way whilst producing qualitatively similar results to the centralised method.

Problems in which there are multiple objective functions to consider, such as the Multi-Objective Vehicle Routing Problem, have seen numerous implementations \cite{Tan2006,Zhang2015,Trivedi2017,Qi2015} of EAs with multiple demes (or populations) in a variety of sizes and structures. Often they use decomposition to break down target problems into smaller sub-problems which are optimized simultaneously and have been shown to be extremely effective at obtaining good solutions.

A number of works look closer at the information exchange and game-theoretical aspects of task allocation \cite{Korsah2013,Kim2011,Walsh1998,Johnson2011,Chapman1980,Shehory1995,Cui2013,Choi2009}. For example the work of Walsh and Wellman \cite{Walsh1998} uses a market based protocol for allocating tasks to agents using a set of bidding policies and auction mechanisms. Alighanbari et. al \cite{Alighanbari2008} develop a robust approach to task assignment for a group of UAVs, where they explore an interesting phenomena that arises when tasks are reallocated too rapidly known as `churning'. This is a type of instability where an agent might start moving towards one task for it to be re-allocated away, requiring them to turn back. Their work looks at ways of anticipating this and hedging against uncertainty to mitigate its impact.

The outline of this paper is as follows, firstly the Multi-Agent Travelling Salesman Problem (MATSP) and the variation of this paper are formulated in \Cref{sec:MATSP}. Next, in \Cref{sec:EATSP}, the Evolutionary Algorithm approach and its applications to the MATSP is outlined. This is initially for an entirely centralised problem, with a single population, then, in \Cref{sec:MDEA}, an implementation of the Multi-Demic Evolutionary Algorithm (MDEA) for solving MATSP is implemented along with the method for decentralising it. Finally in \Cref{sec:results} simulation results for each algorithm for a range of different problem sizes are presented and discussed.

%

\section{MATSP Problem Statement}\label{sec:MATSP}

The Multi-Agent Travelling Salesman, also known as the multiple Travelling Salesman Problem can be formulated in a number of different ways. Here we present the three-index flow-based formulation \cite{Bektas2006a}.

First define the indexes $i$ and $j$ to denote a task from the set $T$ of tasks 1 to $N$, the set $A$ of agents from 1 to $M$ and the matrix $c_{ija}$ to denote the cost of agent $a$ travelling from task $i$ to task $j$. Additionally we define the three-index binary decision variable:
\begin{equation*}
  x_{ija} =
    \begin{cases}
      1 & \text{if agent $a$ visits task $j$ immediately after task $i$,}\\
      0 & \text{otherwise}
    \end{cases}
\end{equation*}
The formulation is then as follows:
\begin{align}
	\min_{x_{ija}} & \sum_{i = 1}^{N}\sum_{j = 1}^{N}\sum_{a = 1}^{M} c_{ija} x_{ija}  \label{MATSP_obj}\\
	\text{s.t.}& \sum_{i = 1}^{N}\sum_{a = 1}^{M} x_{ija} = 1, \text{  } \forall j \label{MATSP_one_task}\\
						 & \sum_{i = 1}^{N}x_{ipa} - \sum_{j = 1}^{N}x_{pja} = 0, \text{  }a \in A, p \in T \label{MATSP_flow} \\
						 & \sum_{j = 1}^{N}x_{1ja} = 1, \text{  } \forall a \in A \label{MATSP_one_agent}\\
						 & u_i - u_j + N \sum_{a = 1}^{M} x_{ija}  \leq N -1, \text{  } \forall i \neq j \neq 1 \label{MATSP_subtour}\\
						 & x_{ija} \in \{0,1\} \text{   } \forall i,j,a
\end{align}
The objective, \Cref{MATSP_obj}, is to minimize the total cost of all the agents travelling between the assigned tasks. The constraints of \Cref{MATSP_one_task} ensure that each task is visited only once while the flow conservation constraints of ~\Cref{MATSP_flow} state that once an agent visits a task then they must also depart from it. The constraints of \Cref{MATSP_one_agent} ensure that each agent is used only once and \Cref{MATSP_subtour} are the subtour elimination constraints \cite{Bektas2006a} where $u$ are additional non-negative auxiliary decision variables, with $u_i$ corresponding to the ith task, known as `node potentials'. Due to the NP hard \cite{Liu2015} nature of the MATSP a direct solution approach may be impractical and likely ill suited for a decentralised implementation. Therefore the approach of this paper is to use the heuristic solution of Evolutionary Algorithms.

A variation to the MATSP is used within this paper, notably the relaxation on the need for agents to start or finish at a depot. This is achieved by representing the agents location as \emph{dummy} tasks, essentially acting as their own personal depot. Along with this an asymmetric extension is made to the cost matrix $c_{ija}$, whereby the cost, $c_{aja}$, is calculated as normal to go from the agent's location (its dummy-task) to each of the other tasks, but the cost to complete the tour, $c_{jaa}$, (i.e. travel from a final task to the dummy agent-task) has a zero cost.
Importantly, in this paper the MATSP is applied to a dynamic simulation of the problem, where the agents progress towards their tasks, complete them and move on to other tasks with new tasks being added over the course of the simulation.

\section{Evolutionary Algorithm for MATSP}\label{sec:EATSP}

\begin{figure}
	\centering
\includegraphics[width=0.9\columnwidth]{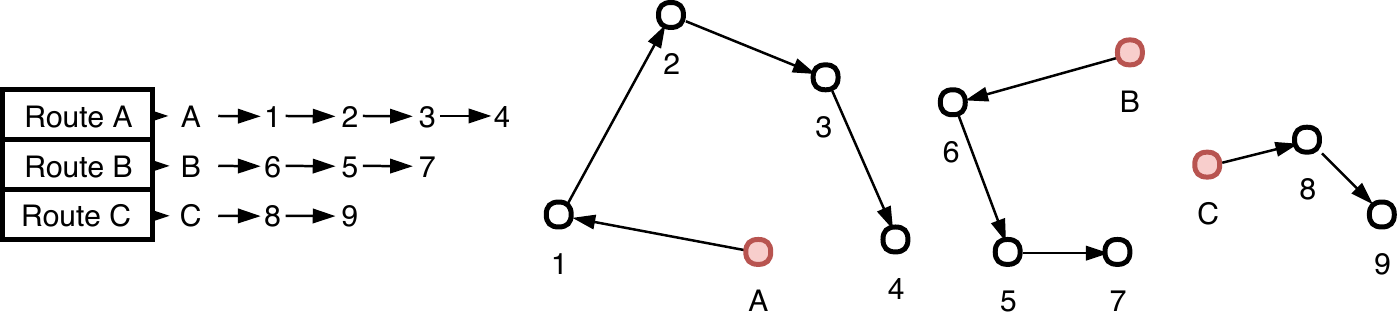}
\caption{Chromosome representation of route-ordered allocation of tasks for three agents}\label{fig_chromosome}
\end{figure}

Evolutionary Algorithms are an increasingly popular heuristic solution method for producing good quality solutions to hard problems in a reasonable time. In order to model the MATSP for use within an EA, first a chromosome representation is developed based on the work of Tan et. al \cite{Tan2006} and depicted in \Cref{fig_chromosome}. More explicitly let us define $T$ to be the set of all $N$ tasks $t_i$ for $i \in \{1..N\}$ and $A$ to be the set of all agents $a \in \{1,..,M\}$. Then let $\tau_k \subseteq T$ be an ordered subset, for each agent $k \in A$, a chromosome $X$, and solution to the MATSP is then defined as:
\begin{gather}
 X:= \{\tau_1, ..., \tau_A \}\nonumber\\
 \text{ s.t } \tau_{a} \cap \tau_{b} = \emptyset, \: \forall \:a \neq b \in A \label{eq_pop_sets}
\end{gather}
A \emph{population}, $P$, is then a set of current chromosomes, $X_l$, defined as
\begin{equation*}
	P := \{X_l\}, \text{ for all } l \in \{1,...,\mu\},
\end{equation*}
where $\mu$ denotes the population size.
Each chromosome in the population also has a `fitness' associated with it, which represents the quality of the individual. As each chromosome describes the routes of all agents then the fitness value is calculated by applying the same path cost summation as \Cref{MATSP_obj}, and thus by finding the individual with the greatest fitness value is equivalent to trying to minimize the MATSP objective of \Cref{sec:MATSP}.

\begin{figure}
\centering
\subfloat[Swap]{\includegraphics[width=0.45\columnwidth]{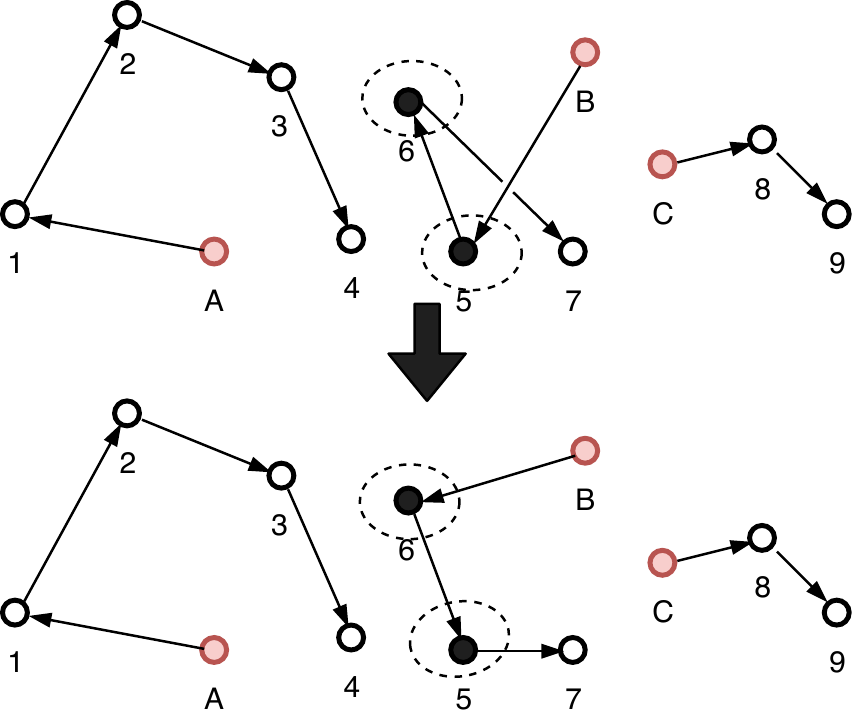}}\hfill
\subfloat[Move]{\includegraphics[width=0.45\columnwidth]{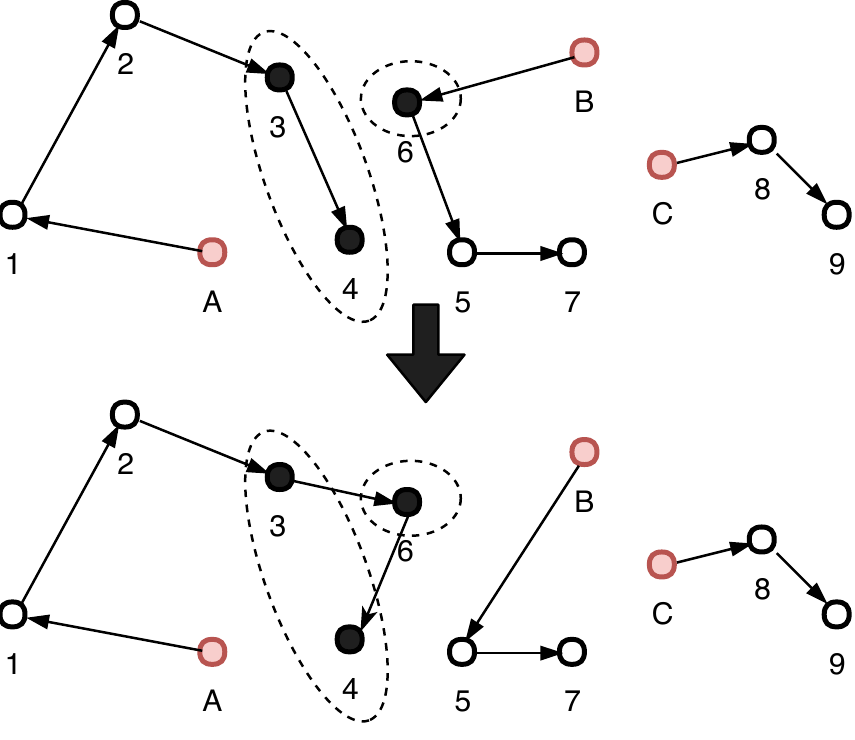}}\hfill
\caption{Mutation strategies before and after}\label{fig_mutations}
\end{figure}

\begin{figure}
	\centering
\includegraphics[width=0.95\columnwidth]{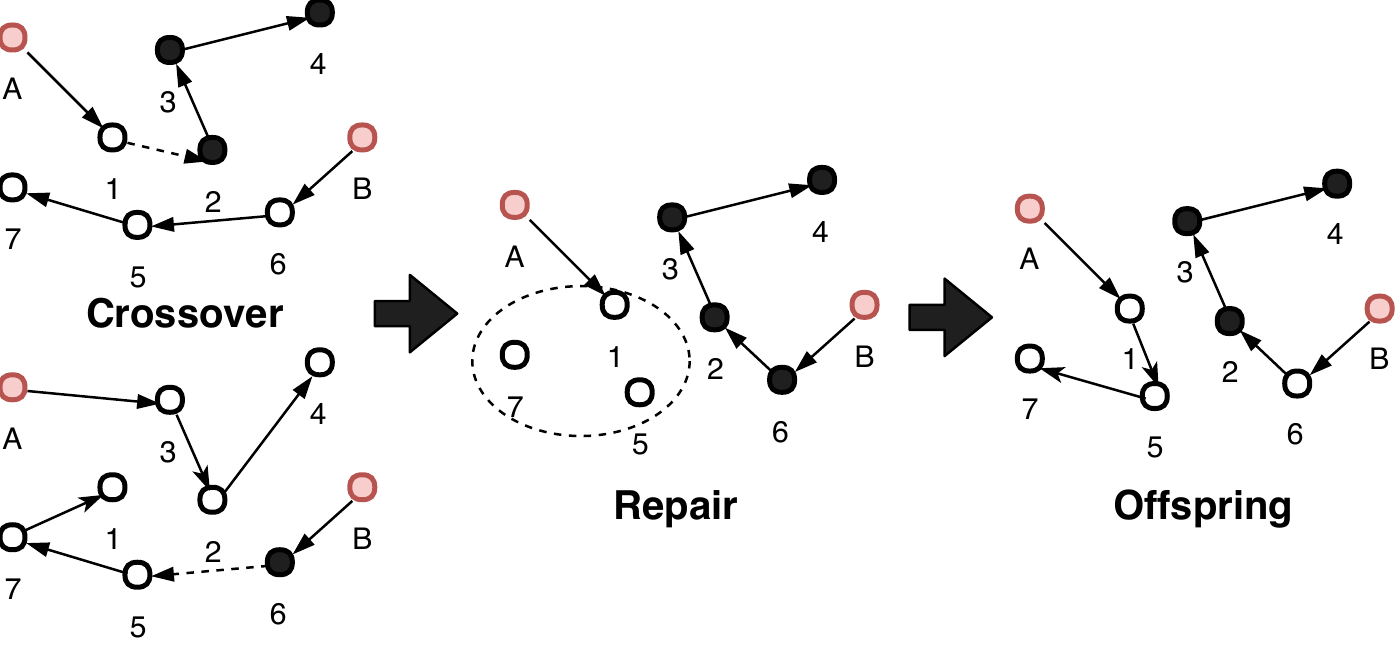}
\caption{Sequence-Based Crossover (SBX)}\label{fig_SBX}
\end{figure}

\begin{figure}
	\centering
\includegraphics[width=0.65\columnwidth]{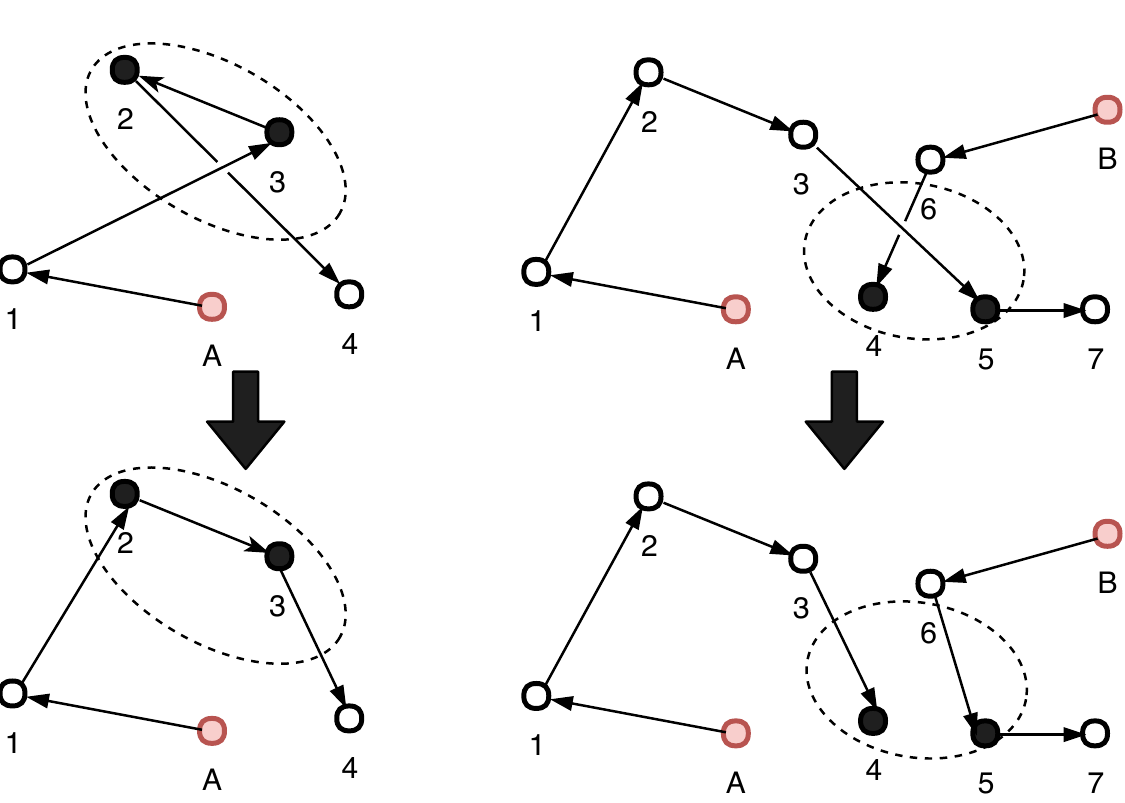}
\caption{Improvement heuristic based on the 2-opt algorithm}\label{fig_2opt}
\end{figure}
An EA can be broken down into 3 main stages:
\begin{enumerate}
	\item \textbf{Initialisation} - creating an starting population for which to evolve;
	\item \textbf{Reproduction} - carrying out evolutionary operators such as crossover, mutation and improvement heuristics to produce offspring;
	\item \textbf{Selection} - taking individuals from both the main population and from the offspring to produce the new population;
\end{enumerate}
The \emph{initialisation} stage is used to create a initial population of feasible solution candidates from which evolution will occur. For the purposes of this paper the initial population was created by assigning each task to the agent which it is closest to with no attempt to optimise the route. The \emph{reproduction} stage is where parents from the main population are used to evolve a number, $\lambda$, of new candidate solutions, or offspring, using three fundamental evolutionary operators, crossover, mutation and improvement. These operators are applied at random, but with fixed probabilities, for a pre-set number of iterations to create a new batch of offspring. It is then in the \emph{selection} stage which looks to combine members of the original population with the newly produced offspring to create a new population, that is, a new generation. The steps of reproduction and selection are then repeated for a number of generations until some stopping criteria is met, in the case of this paper it will be when all tasks have been completed.

There is a large amount of research that focuses on creating evolutionary operators which have favourable properties such as computational efficiency and improved convergence whilst trying to minimise negative properties such as premature convergence and possible undesired speciation \cite{Laumanns2002, Nannen2007, Gomes2015}. The following operators are used specifically for solving a MATSP to utilise the knowledge of the structure of problem to produce better quality candidate solutions and therefore improve performance of the algorithm itself.

\subsection{Mutations}
The two Mutation operators known as the swap-mutation and the move-mutation are based on those implemented by Qi et al. \cite{Qi2015}, these act on a single parent solution and are outlined in \Cref{fig_mutations}. The swap-mutation takes one of the agents at random and two adjacent tasks and swaps their order. Alternatively, the move-mutation takes two agents and moves a random task from one agent route to another.

\subsection{Crossover}
The two crossover operators, Sequence-Based Crossover (SBX) and Route-Based Crossover (RBX) are based on the work of Potvin and Bengio \cite{Potvin} and follow the routine of taking two parent solutions switching parts of the routes and then repairing. The SBX crossover, detailed in \Cref{fig_SBX} takes an agent from each parent, randomly removes a link from each agents route creating a pre-break and post-break route. The pre-break route of one agent is matched with the post-break route of the other to form a new child solution. A second child can be generated in this way by inverting agent from which the pre and post-break routes are taken. Alternatively the RBX crossover takes one agent from each parent and swaps their corresponding route. After fixing any duplicates or unallocated tasks this produces two offspring.
\begin{figure*}
	\centering
\includegraphics[width=1.8\columnwidth]{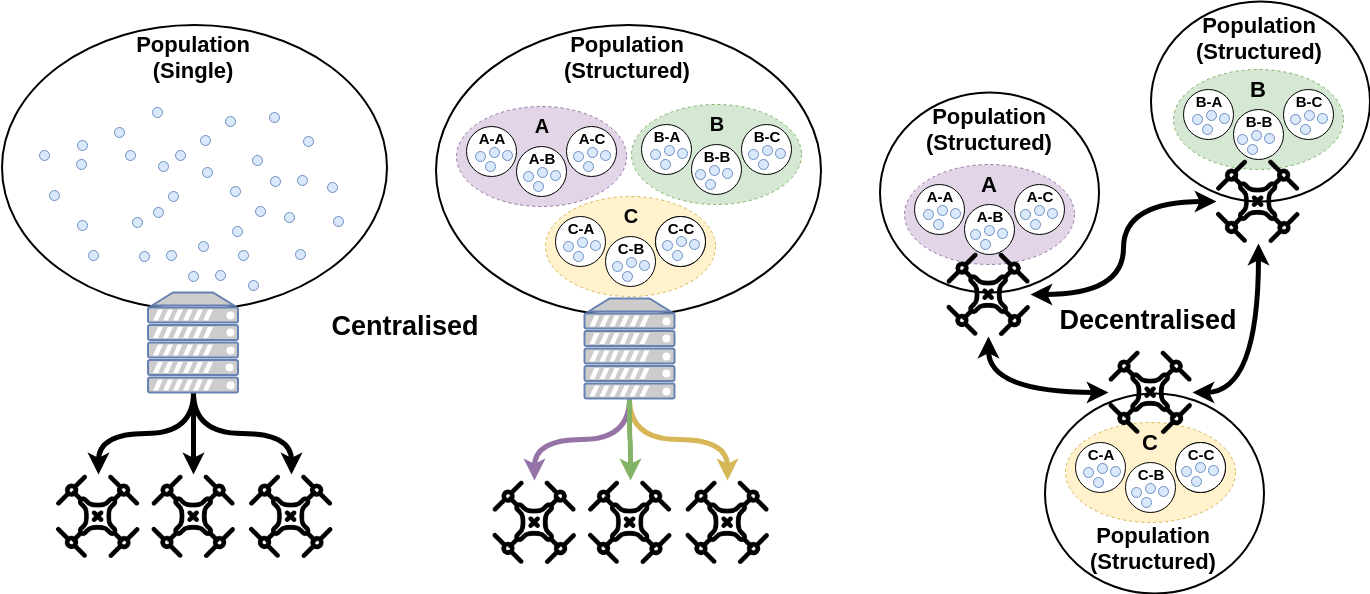}
\caption{Population Structures and exchanges}\label{fig_populations}
\end{figure*}
\subsection{Improvement}
In addition to the two traditional evolutionary operators, mutate and crossover, this work includes an improvement heuristic operator that looks to improve the ordering of agents' routes as shown in \Cref{fig_2opt}. This is done using the the 2-opt method first proposed by Croes \cite{Croes1958} which aims to find routes that cross over themselves and reorder them so they do not. The fact that the underlying problem is a shortest path one allows us to utilise proven methods such as this for making easy improvements to the quality of solution and hopefully improve convergence.



\subsection{Selection}
Selection is an important part of an Evolutionary Algorithm. It determines not only which population members are used for reproduction but also which individuals of the population and new offspring survive until the next generation. This paper uses a random selection method for reproduction and a tournament selection method for determining the new generation of individuals. The tournament method involves running a number of `tournaments' on small batches of the population keeping the winner (the individual with the best fitness). This method has been shown \cite{Alba2002} to be effective as it manages to keep a balance of producing good offspring whilst still allowing `weaker' individuals to propagate in an attempt to improve the search area and avoid premature convergence.

\subsection{Update}
The dynamic nature of simulating the MATSP means that as the simulation runs the state of the world changes and the problem needs to be updated. So, in addition to the evolutionary operators, an update stage is used to move the simulation forward by a time-step $dt$. If the methods of this paper were instead applied to a static problem (i.e. without simulating movement or task addition/removal) then this step would simply be omitted.

It has a number of key tasks:
\begin{enumerate}
	\item \textbf{Move the Agents} - Move the agents towards the next task in their current route (if any);
	\item \textbf{Complete Tasks} - Decide if a task has been successfully completed and set it to complete;
	\item \textbf{Add New Tasks} - Add new tasks to the simulation;
	\item \textbf{Update Distances} - Update the distance matrix for use when evaluating individuals' fitness.
\end{enumerate}
When aspects of communication, which are agent-location dependent, and the addition of new tasks are included, the dynamics of the problem become much more important. The solution to the problem where all tasks are known \emph{a priori} is likely to be very different to one where that knowledge is revealed over time. For the purposes of this paper, as an initial approach, when new tasks are added they are assigned to the geographically closest agent and added to the end of their route. Additionally a newly added task is said to have been completed if the agent is within 1 metre of the task at the time of the update step.

The simulation therefore begins with an \emph{initialisation}, then for every time-step the \emph{reproduction}, then \emph{selection}, then \emph{update} stages are run, with these repeating until all the tasks are completed.


\section{Multi-Demic EA Approach}\label{sec:MDEA}
Large bodies of research focus on managing both the population structure and the individuals that comprise it to improve both the runtime of the algorithm and the quality of solutions it produces. They can typically be classified as either population-distributed, partitioning the individuals of the population, or dimension-distributed, partitioning the problem dimensions. Of the population-distributed, distributed or `island models' and cellular EAs are two of the most popular \cite{Sarma1998,Zhang2015,Tan2006}, relating to the size and spatial structuring of the multiple populations.
The relationship between the structure of the algorithm and populations and the ability to parallelize them has received lots of attention. EAs are fairly amenable to parallelism due to the fact the majority of the operators, such as mutation and selection, can be readily done in parallel. Importantly though, the structuring of the populations leads to the greatest improvements to the algorithm and numerical performance \cite{Alba2002, Gordon1993}.

This paper looks to use the island-model as the EA population-distribution method, where the global population is divided into a number of demes (distinct populations) and referred to as the Multi-Demic Evolutionary Algorithm (MDEA). Communications between these demes allow for individuals to migrate between them at pre-defined intervals. These demes are structured to align with real world execution of a MATSP where tasks are \emph{distributed} amongst multiple agents and are completed independently.

Using the notation of \Cref{sec:EATSP}, each agent, $k \in \{1,..,A\}$, holds a set of demes,
\begin{equation}
	\mathcal{P}_k := \{ P_{kl}\}, \text{ for all }\:l \in \{1,..,A\},
\end{equation}
one deme for each agent $l$ as depicted in \Cref{fig_populations}. Each deme is defined by its parent agent, $k$, and its paired agent, $l$, it is assigned. For each of these demes the evolutionary operators are restricted to alter only parts of the solution which affect the allocation and route of agent $k$ or agent $l$ (the deme's agent pair). Moreover, the demes are used in this way so as to create a situation where each agent has a pairwise way of `reasoning' about potential interactions with the other agent. In this scope it is assumed that each agent has the authority and control over the tasks in which it is allocated, the only way for a task to change allocations (i.e. move to another agent) is for the agent who currently owns the task and the agent the task is moving to, to agree to an exchange. Furthermore each agent $k$ has a `personal' deme $P_{kk}$ which is used to improve its own route without any changes in allocation. In this way, any agent who never interacts with another agent is still capable of optimising its own route and carrying out its allocation.

\begin{figure*}
\centering
\subfloat[Time step 2]{\includegraphics[width=0.9\columnwidth]{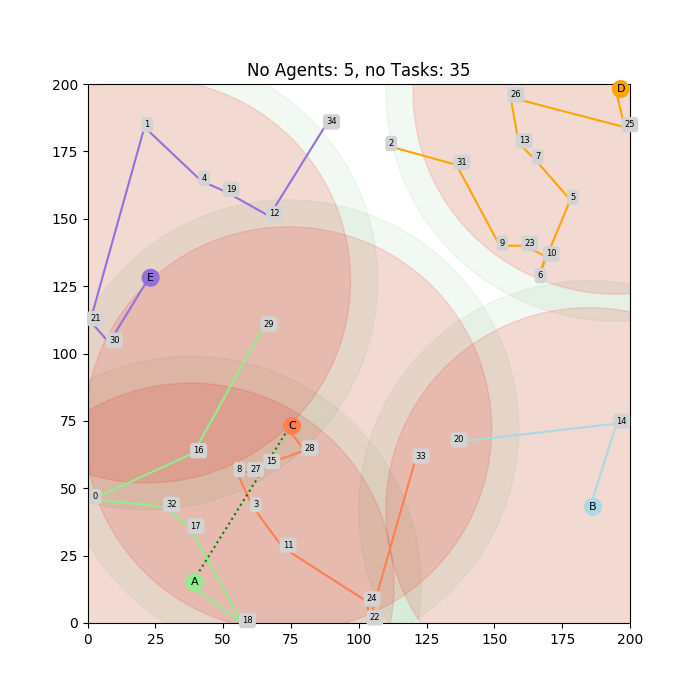}\label{fig:ex_a}}\hfill
\subfloat[Time step 3]{\includegraphics[width=0.9\columnwidth]{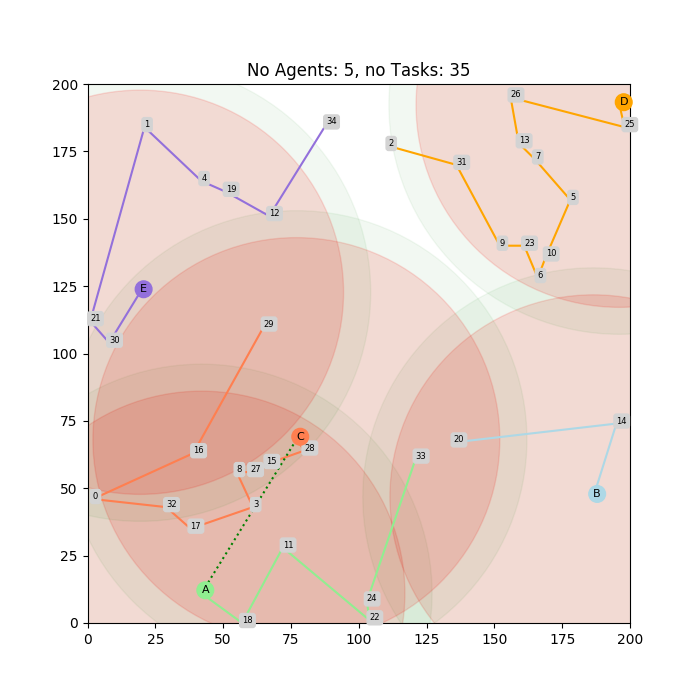}\label{fig:ex_b}}\hfill
\caption{Snapshots of dMDEA solution, with $75$ metre communication (red) and additional consideration radius (green)}\label{fig:ex}
\end{figure*}

\subsection{Exchange}
The MDEA follows a similar evolutionary procedure to the centralised single-population EA, i.e. initialisation, reproduction, selection and update, except these are applied independently to each deme. After a number of generations it is the role of the \emph{exchange} operator to synchronously migrate individuals between demes and to generate the best current route for the agents to follow.
In the centralised EA with a single population the exchange stage's only role is to apply selection to choose the current best individual from the population. This best individual determines the current allocation of tasks and also the route of each agent. In turn the update stage uses this current route to \emph{move} the agents towards their next task.
In the multi-demic scenarios the exchange carries out the following:
\begin{enumerate}
	\item \textbf{Determine feasible exchanges} - Determine which agents are allowed (and are able) to communicate and pass information in preparation for the following steps;
	\item \textbf{1st Knowledge update} - Propagate each agent's current allocation of tasks between agents allowable from step (1) and fix/prune individuals which are invalid;
	\item \textbf{Migrate Compatible Individuals} - Migrate individuals whose allocation is valid for that deme's agent pair between corresponding demes of the other agents i.e $P_{kl} \leftrightarrow P_{lk}$;
	\item \textbf{Exchange Allocations} - In a random order from the list of feasible exchanges check the appropriate demes for the best individual and determine if is worthwhile exchanging tasks.
	\item \textbf{Update Current Best Agent Routes} - Each agent, $k$'s, own allocation and route is determined by selecting the best individual from its personal deme $P_{kk}$;
	\item \textbf{2nd Knowledge update} - Propagate the new allocation of tasks between agents allowable from step (1) and fix/prune individuals which are invalid;
\end{enumerate}

In the case where the MDEA is centralised, and there is no restriction on communication of the feasible exchanges are all possible agent pairings, there is no partial information in the knowledge updates and the best solution can be taken as the best global individual from all demes. In the decentralised case, restrictions on communications are implemented, meaning that agents are only able to exchange when they are within a certain proximity of one another. Here the propagation of knowledge becomes important, agents who have been out of communication range might suddenly find the current state of the problem vastly different from the individuals within its demes. Therefore the two knowledge update steps are key in trying to either repair individuals or prune ones which are completely incompatible.

Structuring the demes and exchanges in such a way as for them to always exchange in a pairwise manner has a number of benefits. Importantly it allows each agent to always be sure that the allocation and route they are following does not conflict with any one else, they are free to evolve their demes and `reason' about potential exchanges with other agents, but if they never come into contact with another agent they are still able to continue on their own. Indeed if no exchanges take place each agent still has its own personal deme for improving on its current route without the help of others.

An example exchange of the dMDEA is shown in \Cref{fig:ex} between time step 2 and 3. Each agent's communication range is shown in red and the extra consideration range in green, what can be seen here is an exchange,  indicated by the dotted green line, between agent $A$ (light green) and agent $C$ (orange) swapping the tasks (17,32,0,16,29) and (11,22,24,33) respectively.

The aim of first distributing the populations and then decentralising the communication is to replicate a likely implementation scenario for a real world problem. In many multi-agent applications, such as search and rescue, agents may have strict communication constraints whilst being geographically spread across large areas. It is likely that an agent may only have an intermittent ability to communicate with a central location. Therefore by decentralising the approach agents are able to have assurance in their current route whilst the system is less sensitive to single-point failures.

A number of potentially important aspects for fully decentralised implementations have not currently been explored in this work. Notably aspects of belief and uncertainty in other agents are neglected so it is assumed the locations of other agents are known and that all information shared is entirely accurate.

\section{Results and Discussion}\label{sec:results}

\begin{figure*}
\centering
\subfloat[3 Agents 25 Tasks]{\includegraphics[width=0.33\textwidth]{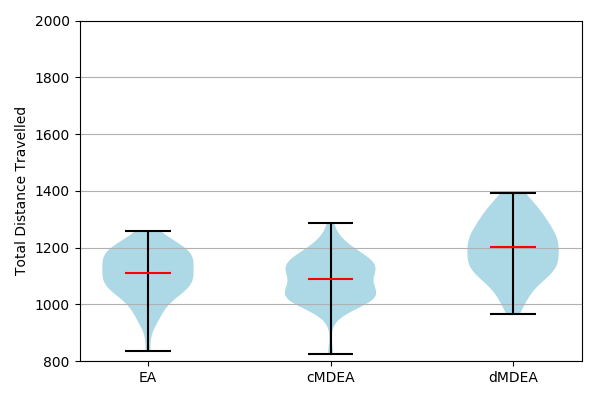}\label{fig:res_total_distance_a}}\hfill
\subfloat[5 Agents 35 Tasks]{\includegraphics[width=0.33\textwidth]{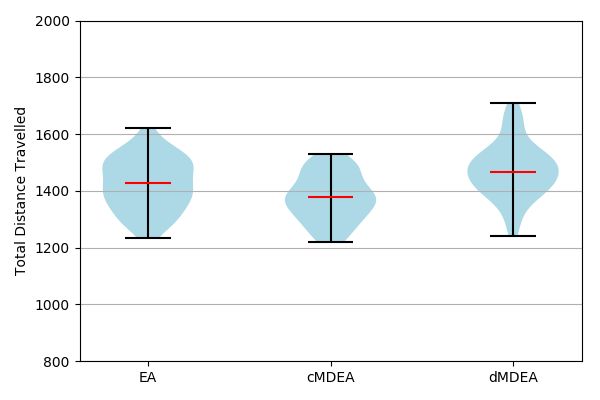}\label{fig:res_total_distance_b}}\hfill
\subfloat[7 Agents 45 Tasks]{\includegraphics[width=0.33\textwidth]{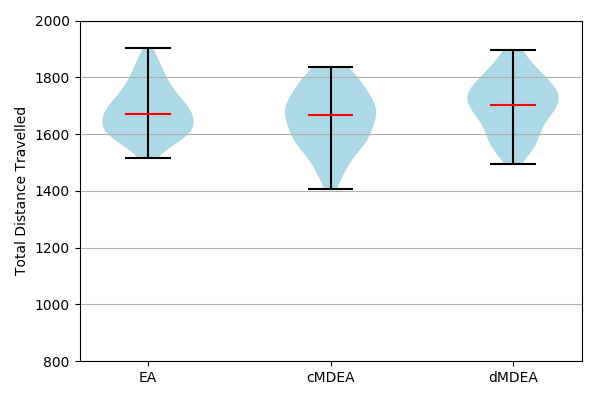}\label{fig:res_total_distance_c}}\hfill
\caption{Total distance travelled}\label{fig:res_total_distance}
\end{figure*}

\begin{figure*}
\centering
\subfloat[3 Agents 25 Tasks]{\includegraphics[width=0.33\textwidth]{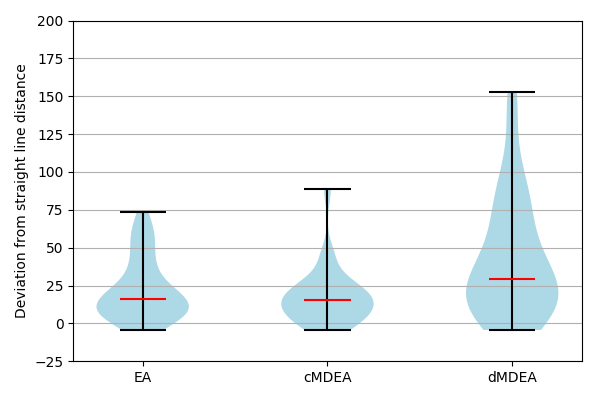}}\hfill
\subfloat[5 Agents 35 Tasks]{\includegraphics[width=0.33\textwidth]{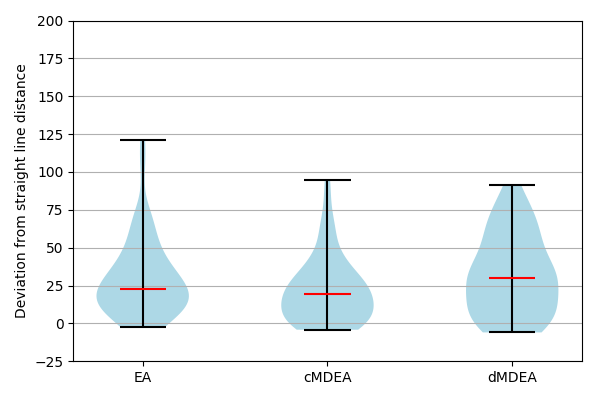}}\hfill
\subfloat[7 Agents 45 Tasks]{\includegraphics[width=0.33\textwidth]{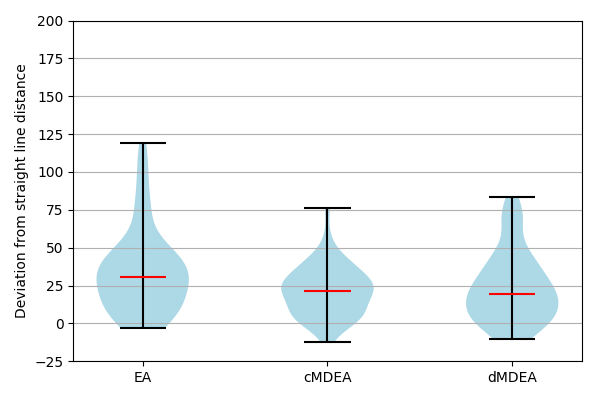}}\hfill
\caption{Deviation from straight line distance}\label{fig:res_dev_straight}
\end{figure*}

\begin{figure*}
\centering
\subfloat[dMDEA communication range vs total distance]{\includegraphics[width=\columnwidth]{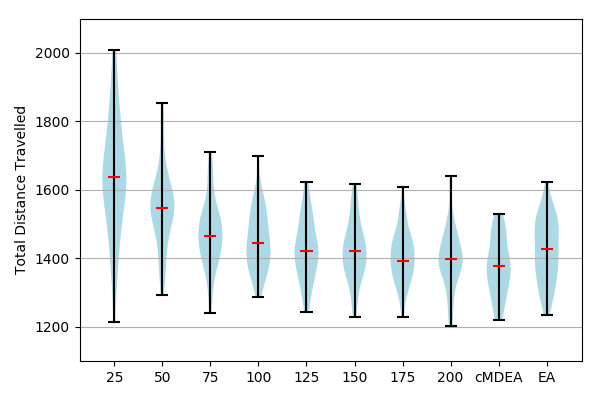}\label{fig:res_radii_a}}\hfill
\subfloat[dMDEA communication range vs run time]{\includegraphics[width=\columnwidth]{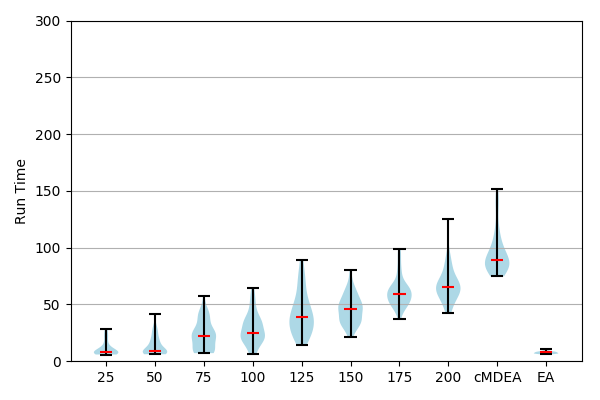}\label{fig:res_radii_c}}\hfill
\caption{Changes in communication distances 5 Agents 35 Tasks}\label{fig:res_radii}
\end{figure*}
Using the outlined methods of \Cref{sec:EATSP,sec:MDEA} we now look to apply the single-population EA, the centralised MDEA (cMDEA) and the decentralised MDEA (dMDEA) approaches to solve a set of sample problems. Let us define a trial as applying a given algorithm (EA, cMDEA or dMDEA) with a given set of parameters (such as number of tasks or number of agents). Additionally a scenario is defined as the initial conditions, such as starting location of the agents and tasks as well as the order and location of addition tasks to arise over the course of the simulation. Given the stochastic nature of Evolutionary Algorithms it is important to try and compare different simulations as fairly as possible. In order to do this each trial is run against the same set of $50$ distinct scenarios and then the spreads of results are analysed. The area in which the problem takes place is a 200 by 200 metre square with agents' initial locations and all tasks being randomly spread with a minimum separation of 1 metre. For each of these trials the initial number of tasks are assigned at the start of the simulation and a further $50\%$ (rounded down) are added one-by-one at set intervals as the simulation progresses. The methods outlined in this paper have been implemented in Python $3.5$ and all the simulations are run on a Dell Precision 3520 laptop running Ubuntu 16.04, with a 2.7Ghz core i7 CPU and 16GB of RAM.

For each scenarios the aim is to minimize the total distance travelled, that is, the objective of \Cref{MATSP_obj} is minimised where $C_{ija}$ reflects this distance cost. In addition all the agents experience this cost function homogeneously at one unit of cost per unit distance. An extensive assessment of the impact of parameters that lead to changes in performance of the evolutionary operators themselves has not be carried out for this work, instead sensible parameters have been chosen that have performed well in testing. The parameters chosen are detailed in \Cref{tab:params}, where it is worth noting that for the multi-demic cases the parameter value is the same for each of the demes. Additionally the fixed probabilities of performing a crossover, mutation or improvement were $40\%$, $40\%$ and $20\%$ respectively, and where those operators had multiple types these were proportioned evenly.

Firstly, the trials have been run for pairs of 3, 5 or 7 agents and 25, 35 or 45 initial tasks respectively, with the additional $50\%$ of tasks added one-by-one every $5$ time steps, with the results plotted in \Cref{fig:res_total_distance,fig:res_dev_straight}. For the dMDEA, each agent has a homogeneous communication range of $75$ metres, and importantly only evolves the demes corresponding to agents $10$ metres beyond that (called a `consideration radius'). Secondly, in order to assess the impact of communication specifically on the dMDEA, trials have been run for varying communications distances from 25 to 200 metres, shown in \Cref{fig:res_radii}.


\begin{table}
  \caption{Evolutionary parameters}\label{tab:params}
  \begin{tabular}{lrrr}
    \toprule
    & EA & cMDEA & dMDEA\\
		\midrule
		Population $\mu$ & 50 & 20 & 20 \\
		Offspring $\lambda$	per generation & 25 & 10 & 10 \\
		Generations per time-step		 & 5  & 5  & 5  \\
  \bottomrule
\end{tabular}
\end{table}

%

\begin{table}
  \caption{Average algorithm run-times (seconds)}\label{tab:runtime}
  \begin{tabular}{lrrr}
    \toprule
   & A=3 N=25 & A=5 N=35 & A=7 N=45 \\
		\midrule
	EA 		&	 6.86 &  7.85 &  10.73\\ 
	cMDEA	& 41.50 &  96.10 &  169.03 \\ 
	dMDEA &	15.12 &  22.97 &  50.39\\ 
  \bottomrule
\end{tabular}
\end{table}
%

The objective function, that is, total distance travelled is the main metric for measuring the performance of each algorithm and is shown in \Cref{fig:res_total_distance}. It can be seen that the cMDEA overall has a positive effect on the outcome of each simulation, with the average total distance travelled being reduced or staying almost identical to the single population EA. However, as noted in \Cref{tab:runtime}, this comes at the cost of a greatly increased linear run-time, which also scales relatively poorly with problem size. This is due to the structuring of multiple demes, for each of the $A$ agents there is a set of $A$ demes, resulting in $A^2$ total demes. This scaling effect also impacts both the number of total population size (over all the demes) and the number of offspring produced per iteration. For the single population EA the scaling of the problem has no impact on the size of the population nor on the number of offspring produced, instead the burden of the problem size is placed on the complexity of solving the MATSP and therefore instead impacts the solution quality. This can be seen in the \Cref{fig:res_total_distance_c}, with cMDEA producing a number of better results, some as much as a $10\%$ improvement.

Even with the restrictions on communication and requirement of decentrality, the dMDEA only really has a significant reduction in performance for the smallest problem size (3 agents and 25 tasks), potentially due to a reduced chance in being within communication range due to fewer agents. As the dMDEA has a communication restriction of $75m$ and the consideration radius of $10m$ beyond that, the agents only evolve the demes corresponding to relatively nearby agents and as a consequence the run-times are greatly improved. Additionally, due to the decentralised nature of the dMDEA it is assumed that the calculations will be done on board each agent and will therefore be done in parallel meaning the wall-time of the system will scale with problem size significantly closer to $\mathcal{O}(A)$ than the $\mathcal{O}(A^2)$ of the cMDEA. Furthermore, only a small minority of the computation time is attributed to the synchronous processes such as exchange and therefore the dMDEA arguably has the potential to be faster that the single population EA.

One interesting dynamic factor, noted by \cite{Alighanbari2008} and coined `churning', is a negative side-effect of frequently altering agents' plan whilst they are en route. To analyse this an measure called `straight line distance' is used. Here the final outcome of the trial is analysed and the straight line distance is summed for each agent between each of the tasks in the order in which they were completed. The idea here is that this is the distance the agent would have taken if this route and order was known ahead of time (note that this is not necessarily the optimal ordering). The measure is then the deviation of the realised path distance against what it could have been and have been plotted in \Cref{fig:res_dev_straight} (Note that the simulations are able to achieve a negative deviation from straight line distance due to the fact a task is considered complete if the agent assigned to it is within 1 metre). Encouragingly, neither the cMDEA nor dMDEA appear to worsen this churning effect even over the three problem sizes.

Aside from decentrality the main difference between the dMDEA and the cMDEA is the restriction on communication distance. To asses the impact of this constraint, for the problem size of $5$ agents and $35$ tasks, the results of the $50$ trials, for communications distances from 25 to 200 metres (and consideration distance of $10m$ beyond that) are shown in \Cref{fig:res_radii}. As you would expect as the communications restriction is gradually lifted the total distances of the dMDEA results tends to the cMDEA. Notably, any communication radius of $125$ or greater either matches or outperforms the EA. What is also clear from \Cref{fig:res_radii_c} is the relationship between the consideration radius (communication radius plus $10$), i.e. the number of other agents to consider, and the run-time. Leading to a possible trade-off decision between ability to communicate, or at least which agents you should consider, and run-time.

\section{Conclusion}
This work has aimed to solve a variation of the dynamic Multi-Agent Travelling Salesman problem, whereby a number of tasks need to be allocated to, and completed by, a number of agents. The focus has been to align the structure of the real world constraints of a problem, such as having geographically diverse agents with limited communication, with the structure of the solution process. Three versions of an Evolutionary Algorithm solution to the MATSP have been proposed. A single population EA a centralised Multi-Demic EA and ultimately a decentralised Multi-Demic EA

Typically adding constraints to a centralised problem results in drops in performance, however the results of \Cref{sec:results} show not only that the cMDEA and dMDEA closely matches the results of the centralised EA but in many cases improves the performance. This improvement is offset by the increase in run-time of the Multi-Demic approaches mostly due to scaling for additional agents. However it is shown that one upside of having a communication restriction in the dMDEA is not needing to constantly consider all other agents and consequently the run-time of the dMDEA is significantly reduced. Moreover, as the dMDEA is decentralised by design, it can be parallelised across each of the agents, actually resulting in faster run-times. For the example of $5$ agents and $35$ tasks it is shown that any communication radius of $125$ or greater either matches or outperforms the EA, thus there is a trade-off between run-time and performance.

There are a number of further research questions that the authors would like to explore. Firstly the impacts of different task assignment schemes such as giving all new tasks to some \emph{leader} agent or only finding them when nearby, perhaps requiring an additional search procedure. Additionally, implementing communication constraints on the single population EA would allow us to more deeply explore the impact of communication on performance of the MATSP. Moreover, it would be of interest to fix the allowed computation time and study the direct impact on performance of varying the other parameters such as population and offspring sizes to fill the time as this might result in fairer comparison. Finally, as robustness is a desirable property often attributed to decentralised problems it would be helpful to be able to directly assess aspects such as the ability for the dMDEA to handle communication drop outs or the loss of an agents whilst still being able to complete all tasks.



\bibliographystyle{unsrt}  
\bibliography{library}  

\begin{thebibliography}{10}

\bibitem{Alighanbari2004}
Mehdi Alighanbari.
\newblock {\em {Task assignment algorithms for teams of UAVs in dynamic
  environments}}.
\newblock PhD thesis, Massachusetts Instittue of Technology, 2004.

\bibitem{Ramchurn2015a}
S.~D. Ramchurn, Joel~E. Fischer, Yuki Ikuno, Feng Wu, Jack Flann, and Antony
  Waldock.
\newblock {A study of human-agent collaboration for multi-UAV task allocation
  in dynamic environments}.
\newblock {\em IJCAI International Joint Conference on Artificial
  Intelligence}, 2015-Janua(Ijcai):1184--1192, 2015.

\bibitem{Groen2007}
Frans C~A Groen, Matthijs T.~J. Spaan, Jelle~R. Kok, and Gregor Pavlin.
\newblock {Real World Multi-agent Systems: Information Sharing, Coordination
  and Planning}.
\newblock In {\em Logic, Language, and Computation}, number 4363, pages
  154--165. 2007.

\bibitem{Korsah2013}
G.~Ayorkor Korsah, Anthony Stentz, and M.~Bernardine Dias.
\newblock {A comprehensive taxonomy for multi-robot task allocation}.
\newblock {\em International Journal of Robotics Research}, 32(12):1495--1512,
  2013.

\bibitem{Bektas2006a}
Tolga Bektas.
\newblock {The multiple traveling salesman problem: An overview of formulations
  and solution procedures}.
\newblock {\em Omega}, 34(3):209--219, 2006.

\bibitem{Alba2002}
Enrique Alba and Marco Tomassini.
\newblock {Parallelism and evolutionary algorithms}.
\newblock {\em IEEE Transactions on Evolutionary Computation}, 6(5):443--462,
  2002.

\bibitem{Zhang2015}
Yue-Jiao Gong, Wei-Neng Chen, Zhi-Hui Zhan, Jun Zhang, Yun Li, Qingfu Zhang,
  and Jing-jing Li.
\newblock {Distributed evolutionary algorithms and their models: A survey of
  the state-of-the-art}.
\newblock {\em Applied Soft Computing}, 34(2013):286--300, sep 2015.

\bibitem{Louis1999}
Sushi~J. Louis, Xiangying Yin, and Zhen~Ya Yuan.
\newblock {Multiple vehicle routing with time windows using genetic
  algorithms}.
\newblock {\em Proceedings of the 1999 Congress on Evolutionary Computation,
  CEC 1999}, 3:1804--1808, 1999.

\bibitem{Lopes2016}
Rui~Borges Lopes, Carlos Ferreira, and Beatriz~Sousa Santos.
\newblock {A simple and effective evolutionary algorithm for the capacitated
  location-routing problem}.
\newblock {\em Computers and Operations Research}, 70:155--162, 2016.

\bibitem{Potvin1996}
Jean-Yves Potvin.
\newblock {Genetic algorithms for the traveling salesman problem}.
\newblock {\em Annals of Operations Research}, 63(3):337--370, jun 1996.

\bibitem{Sarma1998}
Jayshree~A Sarma.
\newblock {\em {An Analysis of Decentralized and Spatially Distributed Genetic
  Algorithms}}.
\newblock PhD thesis, George Mason University, 1998.

\bibitem{Tan2006}
K.~C. Tan, Y.~H. Chew, and L.~H. Lee.
\newblock {A hybrid multiobjective evolutionary algorithm for solving vehicle
  routing problem with time windows}.
\newblock {\em Computational Optimization and Applications}, 34(1):115--151,
  2006.

\bibitem{Trivedi2017}
Anupam Trivedi, Dipti Srinivasan, Krishnendu Sanyal, and Abhiroop Ghosh.
\newblock {A survey of multiobjective evolutionary algorithms based on
  decomposition}.
\newblock {\em IEEE Transactions on Evolutionary Computation}, 21(3):440--462,
  2017.

\bibitem{Qi2015}
Yutao Qi, Zhanting Hou, He~Li, Jianbin Huang, and Xiaodong Li.
\newblock {A decomposition based memetic algorithm for multi-objective vehicle
  routing problem with time windows}.
\newblock {\em Computers and Operations Research}, 62:61--77, 2015.

\bibitem{Kim2011}
Min-Hyuk Kim.
\newblock {Distributed task allocation for multi-robot systems in military
  domain}.
\newblock {\em ProQuest Dissertations and Theses}, 3481052:181, 2011.

\bibitem{Walsh1998}
W.E. Walsh and M.P. Wellman.
\newblock {A market protocol for decentralized task allocation}.
\newblock In {\em Proceedings International Conference on Multi Agent Systems
  (Cat. No.98EX160)}, pages 325--332. IEEE Comput. Soc, 1998.

\bibitem{Johnson2011}
Luke Johnson, Sameera Ponda, Han-Lim Choi, and Jonathan How.
\newblock {Asynchronous Decentralized Task Allocation for Dynamic
  Environments}.
\newblock In {\em Infotech@Aerospace 2011}, St. Louis, Missouri, 2011. American
  Institute of Aeronautics and Astronautics Author's.

\bibitem{Chapman1980}
Archie~C Chapman, Rosa~Anna Micillo, Ramachandra Kota, and Nicholas~R Jennings.
\newblock {Decentralised Dynamic Task Allocation: A Practical Game–Theoretic
  Approach}.
\newblock In {\em Proc. of 8th Int. Conf. on Autonomous Agents and Multiagent
  Systems (AAMAS2009)}, number~8, page~8, 2009.

\bibitem{Shehory1995}
Onn Shehory and Sarit Kraus.
\newblock {Task Allocation via Coalition Formation Among Autonomous Agents}.
\newblock In {\em Proc. of IJCAI}, number 6288, pages 655--661, Montreal,
  Canada, 1995.

\bibitem{Cui2013}
Rongxin Cui, Ji~Guo, and Bo~Gao.
\newblock {Game theory-based negotiation for multiple robots task allocation}.
\newblock {\em Robotica}, 31(6):923--934, 2013.

\bibitem{Choi2009}
Han~Lim Choi, Luc Brunet, and Jonathan~P. How.
\newblock {Consensus-based decentralized auctions for robust task allocation}.
\newblock {\em IEEE Transactions on Robotics}, 25(4):912--926, 2009.

\bibitem{Alighanbari2008}
Mehdi Alighanbari and Jonathan~P. How.
\newblock {A robust approach to the UAV task assignment problem}.
\newblock {\em International Journal of Robust and Nonlinear Control}, 18(2
  SPEC. ISS.):118--134, jan 2008.

\bibitem{Liu2015}
Hong Liu, Peng Zhang, Bin Hu, and Philip Moore.
\newblock {A novel approach to task assignment in a cooperative multi-agent
  design system}.
\newblock {\em Applied Intelligence}, 43(1):162--175, 2015.

\bibitem{Laumanns2002}
Marco Laumanns, Lothar Thiele, Kalyanmoy Deb, and Eckart Zitzler.
\newblock {Combining convergence and diversity in evolutionary multiobjective
  optimization}.
\newblock {\em Evolutionary Computation}, 10(3):263--282, 2002.

\bibitem{Nannen2007}
Volker Nannen and A.~E. Eiben.
\newblock {Relevance estimation and value calibration of evolutionary algorithm
  parameters}.
\newblock In {\em IJCAI International Joint Conference on Artificial
  Intelligence}, pages 975--980, 2007.

\bibitem{Gomes2015}
Jorge Gomes, Pedro Mariano, and Anders~Lyhne Christensen.
\newblock {Cooperative Coevolution of Partially Heterogeneous Multiagent
  Systems}.
\newblock In {\em Proceedings of the 14th International Conference on
  Autonomous Agents and Multiagent Systems (AAMAS 2015)}, number Aamas, pages
  297--305, Istanbul, Turkey, 2015. auton.

\bibitem{Potvin}
Jean-Yves Potvin and Samy Bengio.
\newblock {The Vehicle Routing Problem with Time Windows -Part II: Genetic
  Search}.
\newblock {\em INFORMS journal on Computing}, 8(2):1--21, 1996.

\bibitem{Croes1958}
G.~A. Croes.
\newblock {A Method for Solving Traveling-Salesman Problems}.
\newblock {\em Operations Research}, 6(6):791--812, 1958.

\bibitem{Gordon1993}
V.~Scott Gordon and Darrell Whitley.
\newblock {Serial and Parallel Genetic Algorithms as Function Optimizers}.
\newblock {\em The 5th International Conference on Genetic Algorithms},
  136(1):177--183, 1993.

\end{thebibliography}

\end{document}